\documentclass[sigconf]{aamas}

\usepackage{balance} % for balancing columns on the final page

\usepackage{algorithm}
\usepackage{algpseudocode}
\usepackage{amsmath}
\usepackage{booktabs}       % for \toprule, \midrule, \bottomrule
\usepackage{multirow}       % for \multirow in table headers
\usepackage{colortbl}       % for \rowcolors
\usepackage[table,xcdraw]{xcolor}  % for coloring table rows (gray!10)
\usepackage{tablefootnote}
\usepackage{todonotes}
\usepackage{nicefrac}

\newtheorem{theorem}{Theorem} % No [section], so it's numbered globally

\newtheorem{corollary}{Corollary}[section]

\newtheorem{definition}{Definition}[section]

\newcommand{\E}{\mathsf{E}}
\newcommand{\R}{\mathsf{R}}

\doi{HQII4250}

\makeatletter
\gdef\@copyrightpermission{
  \begin{minipage}{0.2\columnwidth}
   \href{https://creativecommons.org/licenses/by/4.0/}{\includegraphics[width=0.90\textwidth]{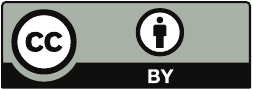}}
  \end{minipage}\hfill
  \begin{minipage}{0.8\columnwidth}
   \href{https://creativecommons.org/licenses/by/4.0/}{This work is licensed under a Creative Commons Attribution International 4.0 License.}
  \end{minipage}
  \vspace{5pt}
}
\makeatother

\setcopyright{ifaamas}
\acmConference[AAMAS '26]{Proc.\@ of the 25th International Conference
on Autonomous Agents and Multiagent Systems (AAMAS 2026)}{May 25 -- 29, 2026}
{Paphos, Cyprus}{C.~Amato, L.~Dennis, V.~Mascardi, J.~Thangarajah (eds.)}
\copyrightyear{2026}
\acmYear{2026}
\acmDOI{}
\acmPrice{}
\acmISBN{}

\title[AAMAS-2026 Formatting Instructions]{A Novel Framework for Uncertainty-Driven Adaptive Exploration}

\author{Leonidas Bakopoulos}
\affiliation{
  \institution{Technical University of Crete}
  \city{Chania}
  \country{Greece}}
\email{lbakopoulos@tuc.gr}

\author{Georgios Chalkiadakis}
\affiliation{
  \institution{Technical University of Crete}
  \city{Chania}
  \country{Greece}}
\email{gchalkiadakis@tuc.gr}

\begin{abstract}
Adaptive exploration methods learn complex policies via alternating between exploration and exploitation. An important question for such methods is to determine the appropriate moment to switch between exploration and exploitation and vice versa. This is critical in domains that require the learning of long and complex sequences of actions. 
In this work, we present a generic adaptive exploration framework that employs uncertainty to address this important issue in a principled manner. 
Our framework includes previous adaptive exploration approaches as special cases. Moreover, it can incorporate any uncertainty-measuring mechanism of choice, such as mechanisms used in intrinsic motivation, or epistemic uncertainty-based exploration methods; and is experimentally shown to give rise to adaptive exploration strategies that outperform standard ones across several environments. Moreover, we showcase its potential for utilization in safety-critical domains. The code for this work can be found at~\url{https://github.com/leoBakop/adaptive_exploration}
\end{abstract}


\newcommand{\BibTeX}{\rm B\kern-.05em{\sc i\kern-.025em b}\kern-.08em\TeX}

\begin{document}

%%% The following commands remove the headers in your paper. For final 
%%% papers, these will be inserted during the pagination process.

\pagestyle{fancy}
\fancyhead{}

%%% The next command prints the information defined in the preamble.

\maketitle 

%%%%%%%%%%%%%%%%%%%%%%%%%%%%%%%%%%%%%%%%%%%%%%%%%%%%%%%%%%%%%%%%%%%%%%%%

\section{Introduction}
Over the past decade, {\em deep reinforcement learning (DRL)}~\cite{Lillicrap2016ddpg,Mnih2015dqn,van2016ddqn} has become the policy learning method of choice in a variety of domains. In online DRL, the goal is to approximate the $Q$-function via alternating between generating new experiences by interacting with the environment (exploration) and updating
the strategy using this experience (learning)~\cite{pislar2022when}.

Now, robotics is considered a standard domain for testing DRL~\cite{fortunato2018noisy,lee2021sunrise,Lillicrap2016ddpg,mahankali2024random}. In general, such domains are characterized by: {\em (i)} continuous and multi-dimensional {\em action} space; {\em (ii)} informative, multi-component, and {\em dense} reward function {\em (iii)} episodes that are very probable to early termination (or ``truncation'') when an agent is set in an unsafe state; {\em (iv)} transitions being highly deterministic. In such domains, the main aim is to learn a complex and long trajectory, necessitating the execution of a sequence of actions that will lead the agent to a specific state. 

In domains with the aforementioned characteristics, it is critical for an agent to pose the question: ``Is it {\em always} better to search for new states, or should I carefully choose {\em when} to explore?'' To further understand the importance of correctly answering that question, let us consider the classic Walker domain~\cite{towers2024gymnasium}. In this domain, a two-dimensional bipedal robot aims to walk in the forward direction. At the beginning of an episode, the agent is set to an initial state. An episode is truncated when the agent finds itself in an unsafe state. Suppose also that the agent has already learned an incomplete and close-to-optimal trajectory which,  in this case, corresponds to performing a single step. This trajectory can guide the agent from the initial state to a new safe one. The agent's goal is to extend this trajectory as much as possible. For that to happen, two phases are required: {\em (i)} the exploitation of the current policy by replicating the trajectory until its end (by replicating in other words the ``single step'' trajectory); and {\em (ii)} the exploration from the end of the trajectory until finding the next best action (explore after performing a ``single step''). We should note that the switch between the two phases should happen within each episode. In case the agent explores instead of replicating the known trajectory, it is more probable to transition to a worse---and maybe unsafe---state rather than the one included in the trajectory. Given this example,~\citet{pislar2022when} argue that it is more advantageous to ask ``{\em when} to explore'' rather than ``{\em how} to explore''.\footnote{The latter is in fact the question that {\em deep exploration} methods~\cite{burda2018rnd,lin2024curse_of_diversity,Lobel2023FlippingCoins,osband2016bdqn,sheikh2022maximizing_ensemble_diversity} focus on. Even though these were designed for use in structurally different domains (e.g., video game domains with sparse, discrete rewards, and stochastic transitions), they are regularly used in robotics as well \cite{Lobel2023FlippingCoins,mahankali2024random}}

In this paper, we present {\em ADEU - ADaptive Exploration via Uncertainty}, a generic exploration framework that decides in a {\em principled} manner {\em when} the switching between exploration and exploitation and vice versa should take place within an episode.
Intuitively, ADEU enables exploration when it adjudges that its {\em uncertainty} regarding its policy is ``high enough''. Now, uncertainty is arguably a subjective term and is highly dependent on the formulation of the problem being solved. ADEU takes that into account. It can, in fact, incorporate {\em any} mechanism for measuring uncertainty in its action selection mechanism. Intuitively, such an {\em uncertainty-measuring} mechanism corresponds to a particular ``uncertainty definition''. Moreover, ADEU does not resort to the use of heuristic techniques or thresholds to answer the aforementioned {\em ``is uncertainty high enough''} question: instead, it selects an action in a principled manner, by sampling from a distribution whose spread is proportional to its uncertainty regarding the policy. By being able to incorporate any uncertainty-measuring mechanism, and by selecting an action in the aforementioned principled manner, ADEU is, in fact, able to generalize over existing adaptive exploration strategies. Furthermore, we prove in the paper that ADEU can achieve optimal exploratory behaviour for a wide class of games. Finally, we put forward ADEU versions that can {\em safely explore} in safety-critical environments,
as confirmed by preliminary results.

To test our framework, we create different instances of ADEU, each corresponding to a different uncertainty-measuring mechanism.
We test our method in the most complex/difficult among the robotic tasks in~\cite{towers2024gymnasium}, and in the DeepSea domain.
%---a domain which tests the agent's ability for deep exploration. 
Our experiments demonstrate that the performance of an ADEU instance depends on the ability of each uncertainty-measuring mechanism to quantify uncertainty.  
However, in our experiments, the instance of ADEU employing a specific uncertainty-measuring mechanism, outperforms the original exploration method that employs the same uncertainty-measuring mechanism. 
Moreover, there is {\em always} some instance of ADEU that performs better than all of its opponents.
As such, ADEU can be considered as a `plug-and-play' adaptive exploration framework, that can be coupled with any uncertainty-measuring mechanism; and exhibits the ability to improve the performance of DRL algorithms that utilize the corresponding uncertainty-measuring mechanism in their exploration process.

\section{The ADEU framework}
\begin{figure}[hbtp]
    \centering
    \includegraphics[width=0.48\linewidth]
    {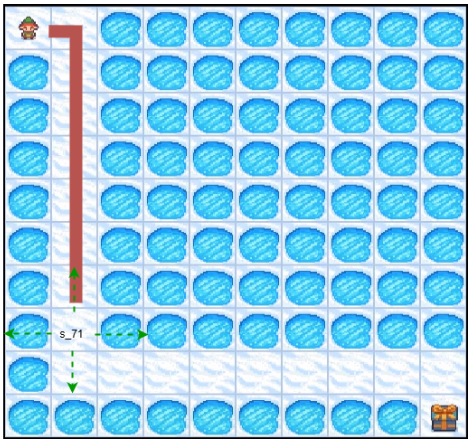}
    \caption{The modified Frozen Lake environment. Exploration gets harder as dimensions increase.}
    \label{fig:motivation_example}
\end{figure}

\paragraph{A Motivating Example:}
Let us consider a variation of the classic Frozen Lake example as shown in Figure~\ref{fig:motivation_example}.\footnote{Frozen Lake's official documentation:~\url{https://gymnasium.farama.org/environments/toy_text/frozen_lake }} In that environment, the agent has to discover the sole trajectory that will eventually lead it to the goal, passing through frozen tiles, while if it steps on a ``hole'' tile the episode terminates. In our example, the agent is rewarded when it reaches the next frozen tile. 
This reward is inversely proportional to the distance of the goal: the reward function outputs a higher reward when the agent reaches an allowed tile that is also closer to the goal. In addition, let us assume that an agent has already been trained to reach state $61$ (state at the end of the red line in Figure~\ref{fig:motivation_example}---right above state $71$). In other words, its policy $\pi(s)$ can successfully lead it to state $61$, replicating the red trajectory. The agent aims to extend the red trajectory until reaching the treasure. 

An effective way for that agent to explore this environment is by ``trusting'' its policy until state $61$ and then switching to exploration. Assume an agent employing a classic exploration technique like $\epsilon$-greedy (or any noise-based strategy) that explores every episode with a {\em constant} $\epsilon$ (or constant variance in the case of a noise-based strategy).  By not modifying $\epsilon$ during an episode, the agent explores at a constant rate. Then, the agent will probably not be able to replicate the red trajectory since it is probable (approximately $\epsilon$\% at each timestep) that the agent will deviate from the known policy.

In an even worse scenario, assume an exploration strategy that {\em forces} the agent to visit states with high uncertainty (e.g., a state that has not been visited enough times, or a state whose value the agent is unsure about). In that scenario, the agent is uncertain for all the states not belonging to the trajectory. Thus, this exploration strategy will force the agent to needlessly explore rather than trusting the trajectory, leading to the truncation of a large number of episodes before managing to actually extend the trajectory. 
The use of a {\em greedy-in-the-limit} exploration strategy would lead to similar ``over-exploration'' phenomena.

Now, suppose an adaptive exploration method that determines at every step whether {\em (i)} the agent should explore a state, or {\em (ii)} it should exploit/trust the known policy for that state. By appropriately alternating between these two phases, the agent can replicate the known trajectory and subsequently begin exploration. A key challenge lies in determining how the agent identifies {\em the appropriate moment} to transition between the phases.

Let us define $f(s) = \frac {1}{\beta \sqrt{n(s)}}$ where $n(s)$ is a (pseudo-)counter that counts (or approximates in larger environments) the times that a state $s$ has already been visited, and $\beta<1$ a positive constant. Let us also suppose that the states of the trajectory have been visited numerous times. Then, $f(s)$ arguably provides an effective signal on whether to switch/alternate between the two phases. This suggests an agent should explore proportionally to the value of $f(s)$.

\subsection{Formal Framework Description}\label{sec:our_method}

Against this background, we put forward ADEU, a framework allowing for adaptive exploration, based on the uncertainty associated with each state. ADEU 
makes use of some uncertainty-measuring mechanism $f(s)$,
%In other words, the value of $f(s)$
whose value dictates when an agent should start exploring and when it should trust the gained knowledge, similarly to the example. As discussed earlier, uncertainty is a subjective term; its definition is closely related to the problem being solved. The ADEU framework implicitly accepts any definition of uncertainty via the implementation of an $f(s)$. As such, practically all uncertainty-measuring mechanisms---e.g., those utilized by {\em intrinsic motivation} methods to detect novel states, or by {\em epistemic uncertainty}-driven exploration to quantify the convergence degree of a policy---are compatible with ADEU.
We will elaborate on this after formally presenting the key ADEU principles.

The ADEU framework can be summarized concretely by Equation~\ref{eq:adeu}, which defines the way that an ADEU agent selects an action given a state $s$. 

\begin{equation}\label{eq:adeu}
    a \sim \mathcal{D}\left (\pi(s), g(f(s))\right )
\end{equation}
where $\mathcal{D}$ is the distribution of choice, $\pi(s)$ is the current policy for a state $s$, $f(s)$ is the aforementioned uncertainty-measuring mechanism and $g()$ is a normalizer function. 

In more detail, this action will be chosen by sampling from a distribution $D$. The first moment of that distribution is the policy $\pi(s)$ that the DRL agent has already learned. Hence, the {\em mean} of the probability distribution $D$ (intuitively, its center of mass) is the policy $\pi(s)$. 
%, and the sampled actions are close to this policy. 
The {\em second moment (variance)} of that 
distribution is $g(f(s))$,
which is proportional to the uncertainty. 
Hence, $g(f(s))$ defines how ``spread'' the distribution is, and how close the sampled action $a$ will be to the policy $\pi(s)$.

The key intuition in ADEU is that the spread of the distribution should be defined so that it corresponds to the agent's uncertainty regarding the $\pi(s)$ policy; while that uncertainty can be calculated in any way considered appropriate.\footnote{In this paper, we use the first raw moment (mean) and the second central moment (variance), both in the context of continuous and discrete distributions. One could conceivably use any other moments of choice to enhance action selection. For instance, one could use some function of the third standardized moment in order to guide exploration towards a specific direction, e.g., a direction that is deemed {\em safe}~\cite{wachi2023safe,ijcai2022p3o}, if such a direction can be in some way associated with the kurtosis of the distribution.}
To this end, $f(s)$ in $g(f(s))$ is {\em some} mechanism able to compute uncertainty, like the ones used in uncertainty-based exploration methods\cite{Chen2017ucb,lee2021sunrise,osband2016bdqn} or mechanisms designed for intrinsic reward construction in intrinsic motivation methods\cite{Badia2020NeverGiveUp,burda2018rnd,Lobel2023FlippingCoins}.\footnote{To clarify, in our method, we use the mechanism used in intrinsic motivation methods for defining $f(s)$, and {\em not} for providing an exploration bonus.} 
By default, $f(s)$ will output a positive scalar. In multidimensional distributions $D$, this scalar can multiply an Identity matrix. 
In addition, since $f(s)$ can be {\em any} uncertainty-measuring mechanism, we use $g(\cdot)$ as a normalizer to prevent $f(s)$ from outputting enormous values. Equipped with $f$ and $g$, ADEU can effectively utilize any uncertainty-measuring mechanism proposed in previous (deep exploration) works.

Moreover, $D$ can be {\em any} distribution deemed appropriate. In this paper, we use the Multinomial distribution in discrete action space experiments/examples, and the Gaussian (or Normal) distribution in continuous action space ones. We selected these distributions due to their simple implementation.

Guided by Equation~\ref{eq:adeu} to select actions, an ADEU agent is able to alternate between exploitation and exploration within the episode in a principled manner, and its behaviour can be summarized as follows. At every timestep, the agent {\em (i)} calculates the uncertainty  $f(s)$ regarding that state;  {\em (ii)} constructs a distribution $D$ in which the first moment is the current policy $\pi(s)$, and the second moment is the $g(f(s))$; and {\em (iii)} samples from that distribution. In case of low uncertainty, the distribution will not be ``spread'' and hence the sampled action will be close to $\pi(s)$. Notice that this effectively causes the agent to replicate parts of a trajectory corresponding to a $\pi$ it is confident about. In any other case, the sampled action will probably be far from $\pi(s)$, leading the agent to explore new actions in that state. This new experience can potentially lead to an extended trajectory.

Note that the discussion above does not imply that ADEU needs any trajectories or other forms of background knowledge.
ADEU assumes a zero-length initial trajectory at the beginning of the training process, and then is progressively able to extend it via its adaptive exploration technique.\footnote{Of course, if ADEU is provided with an accurate initial trajectory or policy, it can exploit such knowledge.} 
In other words, at the start of training, the agent has not yet converged at a policy, even for the initial states. 
%In other words, 
As such, it lacks certainty about which actions to take and must explore until it converges to a reliable policy, even for the initial state. Once this occurs, the agent follows the learnt action policy from the initial state until it reaches the next state where ``enough'' uncertainty arises. There, it resumes exploration. This exploration and policy refinement cycle continues until the agent becomes confident across the entire trajectory. 

We now proceed to define a generic class of exploration-related decision problems (or exploration-related single-agent ``games'')  such as the ``modified Frozen Lake'' one, for which ADEU agents are able to achieve optimal exploratory behaviour. These are``increasing-reward'' settings: intuitively, only the best action transfers the agent from state $s$ to a ``better'' state $s^*$ where $\max_ar(s^*,a) \geq \max_ar(s,a)$. (In our definition and discussion below, $s$ is the current state, $s'$ some future ones, and $s^*$ is the best future one. We will also say that $s$ ``neighbours''  $s'$ if $T(s, \cdot, s') > 0$).

\begin{definition}[Increasing-Reward Single-Agent Games]\label{def:game}
    In {\em Increasing-Reward Single-Agent Games}, an agent has to decide on an action $a$ to execute at a state $s$, given that the following hold. 
    \begin{itemize}
        \item $ \exists ! \langle s^*, a^* \rangle: r(s^*,a^*) \geq \max_ar(s,a)$. There is a unique future $s^*, a^*$ pair, with $s^*$ neighbouring $s$, in which the agent achieves a higher reward than when playing $\arg\max_ar(s,a)$ in $s$.
        \item $ \forall s' \neq s^*,\quad \exists a : max_{a'}r(s',a') < r(s,a)$. That is, for {\em all} future states $s' \neq s^*$ ``neighbouring'' $s$, the agent achieves a lower maximum reward than the reward achieved in the current state $s$ for at least one action $a$.
        \item $\exists! b: b =  \arg\max_ar(s,a), \text{ }\forall s $. That is, at every state $s$, there is a unique action $b$ that maximizes the reward function in that state $s$. Moreover, for that same $b$ it holds:
        \item $T(s, b, s^*) = 1$, where $r(s^*, a^*) > \max_{a'} r(s', a')$ where $s'$ neighbours $s$, $s'\neq s^*$, and $\langle s^*, a^* \rangle$ is the unique pair mentioned above---i.e., when the agent executes $b$ in $s$, it will transition with probability $1$ to the (unique) $s^*$ that is (strictly) better (in terms of maximum reward) than any other neighbour state $s'$.
        %Note that $T$ is the standard transition function~\cite{Sutton1998RL}.
        \item Increasing-Reward Single Agent Games inherit all other 
        %non-reward-function-related
        characteristics from standard MDP models~\cite{Sutton1998RL}. 
    \end{itemize}
\end{definition}

\begin{corollary}
 In Increasing-Reward Single-Agent Games, the optimal behaviour for an agent is to explore a state $s$ until action $b: b = \arg\max_ar(s,a)$ is found; no further exploration is then required for $s$.
 %that state. 
\end{corollary}

Notice that a plethora of standard games (such as the one described in Fig.~\ref{fig:motivation_example}, and also sorting, finding the sole existing trajectory in complex maze environments, etc.) are considered as Increasing-Reward Single-Agent Games. We now prove that ADEU (equipped with an effective $f(s)$) achieves 
efficient
exploratory behaviour in such games with a large but finite action space. In more detail, assuming that after a finite amount of time, the ADEU agent will discover action $b$ for a state $s$, the $f(s)$ will decrease and stop the exploration in that specific state $s$. Hence, for this ADEU agent $\pi(s) = b$ and the agent will explore the future state $s^*$. A suitable $f(s)$ should measure uncertainty as, e.g., the difference between the current policy $\pi(s)$ and $b = \arg\max_ar(s,a)$; or as the $f(s)$ used in the modified Frozen Lake 
%Only for the arXiv version
%(the latter is described in the Appendix).
%%%%%%%%%%%%%%%%%%%%%%%%%%%%
\begin{definition}\label{def:best_states}
    Following the aforementioned notation, we define as ``best neighbour state at timestep $i-1$''  the state $s_i^*: T(s_{i-1},b,s_i^*) = 1$ and $b = \arg \max_a r(s_{i-1},a)$
\end{definition}

\begin{theorem}\label{theorem}
    %Let $s_i^* \text{, where } i=1,...$ states that satisfy Definition~\ref{def:best_states}. Similarly, $b_i = \arg\max_ar(s_i, a)$.
    %ADEU agents provided with an effective uncertainty-measurement mechanism $f(s)$, and
    ADEU agents that have executed  
    $b: = \arg\max_ar(s_i,a)$ for some $s_i$
    during a training episode, 
    will reduce their exploration of $s_i$ relative to the effectiveness of their $f(s)$ mechanism and focus on exploring $s_{i+1}^*$.
\end{theorem}
\begin{proof}
The proof can be found in the Appendix.
\end{proof}
Theorem~\ref{theorem} shows that an ADEU agent, when provided with an effective uncertainty-measuring mechanism, will solve an Increasing-Reward Single-Agent Game. The theorem can be extended to guarantee efficient (in expectation) ADEU behavior in environments with stochastic reward or transition functions also.
%%%%%%%%%%%%%%%%%%%%%%%%%%%%%%%%%%%%%%%%%%%%%%%%%%%%%%
%The following will be added only in the arXiv version
%For extended Theoretical results, please refer to the Appendix.
%%%%%%%%%
ADEU's efficiency arises from the fact that it does not needlessly explore, but instead replicates the parts of a  ``known'' or ``learnt'' trajectory that it trusts. But this trajectory might be sub-optimal. To keep improving the learnt trajectory, ADEU also performs the so-called rollout episodes. In more detail, at the beginning of every episode, with probability $\rho$, ADEU performs exploration using as uncertainty mechanism the $f_\text{rollout}(s) = c$, where $c$ is a constant number. %In other words, at the beginning of an episode, the ADEU agent replaces $f(s)$ with $f_\text{rollout}(s) = c$ with probability $\rho$. 
Hence, for that {\em specific} ``rollout'' episode, ADEU takes the chance to explore so as to improve the learnt trajectory. We should note that in the case of a well-constructed $f(s)$ that smoothly decreases, 
%the uncertainty, 
$\rho$ should and can simply be set to $0$. 
For simplicity, in our implementation we treat $\rho$ as a constant 
that does not change 
throughout 
%the 
training. 
%process. 
Also, we set $f_\text{rollout}(s) = c$ to present a simple version of our framework, but, in more complex environments, this can change too. %We elaborate on this in the Appendix.

Both the rollout episodes and the potential use of more complex definitions of uncertainty, allow the ADEU agent to effectively explore different, and more complex settings (e.g., even settings requiring deep exploration like DeepSea~\cite{osband2019deepSea}) than the ones in Definition~\ref{def:game}, as discussed immediately below and as our experiments in Section~\ref{sec:experiments} demonstrate.

\subsection{Generality of the ADEU framework}\label{subsec:generality}

In this section, we elaborate on the fact that ADEU is a generic adaptive exploration framework. 
First, ADEU can readily use as $f(s)$ any predefined mechanism---e.g., some epistemic uncertainty-measuring mechanism or mechanisms used in intrinsic motivation methods to approximate visitation frequency.
Moreover, it is a framework that can emulate most, if not all, existing adaptive exploration techniques. This can be achieved by designing an as-complex-as-required $f(s, \cdot)$--- 
e.g., an $f(s, \cdot)$ mechanism that outputs a value considering previous states or actions, or other constraints.

\paragraph{Using predefined mechanisms as $f(s)$:} 

%We now elaborate on the fact that ADEU is a generic adaptive exploration framework. In particular, it is a framework that can readily incorporate {\em (a)} exploration strategies belonging in large families of uncertainty-based exploration methods; as well as {\em (b)} existing adaptive exploration techniques.

To see this, consider first the {\em intrinsic motivation (IM)} family, which rewards the agent whenever a novel state is visited. This non-stationary reward, termed as $r_\text{intrinsic}(s)$, is added to the external reward provided by the environment, as described in Equation~\ref{eq:im-original}. Then, the agent acts greedily to maximize~\ref{eq:im-original}.
\begin{equation}
\label{eq:im-original}
r(s) = r_\text{extrinsic}(s) + r_\text{intrinsic}(s)
\end{equation}
To calculate the $r_\text{intrinsic}(s)$, IM-oriented methods~\cite{Badia2020NeverGiveUp,burda2018rnd,Lobel2023FlippingCoins,mahankali2024random} propose numerous mechanisms that can quantify the novelty of a state. For example, RND~\cite{burda2018rnd} proposes to calculate the visitation frequency as the approximation error between two networks named Target and Predictor. Predictor network is trained to predict the Target's output for a given state. The difference between the output of the two networks (Equation~\ref{eq:rnd}) approximates the visitation frequency for that specific state. 
\begin{equation}\label{eq:rnd}
    r^\text{RND}_\text{intrinsic}(s) = \left\lVert f_{\text{predictor}}(s) - f_{\text{target}}(s) \right\rVert_2^2
\end{equation}

ADEU can utilize this approximation error as an uncertainty-measuring mechanism. That is, an instance of ADEU that defines uncertainty as the visitation frequency of a state can utilize $f(s) = r^\text{RND}_\text{intrinsic}(s)$. In a similar way, any other mechanism approximating visitation frequency, such as the ones proposed in other IM methods~\cite{Badia2020NeverGiveUp, Lobel2023FlippingCoins,tang2017exploration} can also be employed. 

One can also create an instance of ADEU that utilizes an uncertainty mechanism similar to the ones defined in epistemic-uncertainty driven methods~\cite{Chen2017ucb,lee2021sunrise}. {\em Epistemic uncertainty}-driven methods employ different representations of the same Q-function to measure how certain an agent is about its policy in a specific state. For example, the UCB~\cite{Chen2017ucb,lee2021sunrise} strategy greedily selects actions that maximize the upper bound of the $Q$-function shown in Equation~\ref{eq:ucb}.
\begin{equation}\label{eq:ucb}
a = \arg\max_{a} \left[ Q_\text{mean}(s, a) + \lambda Q_\text{std}(s,a) \right]
\end{equation}
where $Q_\text{mean}(s, a)$ is the mean $Q$-value calculated across the different representations of the $Q$-function, $Q_\text{std}(s,a)$ is the corresponding standard deviation, and $\lambda$ is a positive constant.
Given this, we can also create an ADEU instance that utilizes the above uncertainty-measuring mechanism by setting  $f(s) =\lambda Q_\text{std}(s, \pi(s))$. Table~\ref{tab:simple_adeu_instances} 
summarizes the discussion above.
%shows these different definitions of uncertainty and how they can be used in different instances of ADEU. 

\begin{table}[h!]
\centering
\caption{\label{tab:simple_adeu_instances}Instances of ADEU using different uncertainty-measuring mechanisms---examples only; {\em any} IM or mechanism calculating confidence over $\pi$ can be used.}
\resizebox{0.6\linewidth}{!}{%
\begin{tabular}{|p{2.5cm}|p{5.3cm}|}
\hline
\textbf{Definition of Uncertainty} & An {\em example} of $f(s)$ \\
\hline
Novelty of $s$ & $f_\text{RND}(s) = \left\lVert f_{\text{predictor}}(s) - f_{\text{target}}(s) \right\rVert_2^2$  \\
\hline
Confidence on $\pi(s)$ & $f_\text{UCB}(s) =  \lambda Q_\text{std}(s, \pi(s))$\\
\hline
\end{tabular}
}
\end{table}

\paragraph{Emulating adaptive exploration techniques:} 

Table~\ref{tab:previous_works} presents certain key adaptive exploration works, and their instantiation within the ADEU framework. See also ``Related Work'' for more details.

\begin{table}[h!]
\centering
\caption{\label{tab:previous_works}
Previous adaptive exploration methods as special cases of ADEU.}
\resizebox{1\linewidth}{!}{%
\renewcommand{\arraystretch}{1}
\begin{tabular}{|p{5cm}|p{5cm}|}
\hline
\textbf{Description} & \textbf{Special Case of ADEU} \\
\hline
\citet{tokic2010adaptive}: $\epsilon$-greedy where $\epsilon$ is adapted at each state based on TD error. & Utilizing a Bernoulli distribution as $D$. Let $f(s)$ = TD-Error$(s)$. \\
\hline
\citet{dabney2021ez_greedy}: Executes a fixed-length option with prob. $\epsilon$; otherwise follows $\pi(s)$. Length sampled from a heavy-tailed distribution. & A Bernoulli decides between $\pi(s)$ and an option. We will utilize a more complex uncertainty-measuring mechanism $f(s,n)$, that calculates uncertainty according (also) to a parameter $n$---this is still allowed by ADEU as previously discussed. This mechanism sets the probabilities $p$ of the Bernoulli based on $n$, which is sampled once and decremented at each step.\tablefootnote{$n$ is sampled in the first call to $f(s,n)$, then
%passed and
decremented in subsequent calls.} \\
\hline
\citet{pislar2022when}: Explores intra-episodically after a trigger signal, for a fixed duration. & $f(s)$ replaces the trigger. A Bernoulli distribution is utilized. $f(s)$ modifies the probability $p$ to favor exploration, then outputs a constant $p$ for a fixed number of steps. \\
\hline
\citet{ecoffet2019goexplore,Ecoffet2021first_return_then_explore,torne2023breadcrumbs}: The agent first returns to a state, then explores. & Use Bernoulli or Gaussian. $f(s)$ is binary and switches phases. \\
\hline
\end{tabular}
}
\end{table}
%\new{
Table~\ref{tab:previous_works} explains how ADEU incorporates previous works on adaptive exploration. For instance, ADEU can emulate ~\citet{dabney2021ez_greedy} by utilizing a Bernoulli distribution, which will ``decide'' over two different scenarios: {\em (i)} policy $\pi(s)$ with probability $1-p$; and {\em (ii)} an option with probability $p$. The $p$ parameter of the Bernoulli distribution will dynamically change according to $f(s,n)$. In more detail, the initial value for $n$ is $0$. At the first call of $f(s,n)$, $n$ will be sampled from a heavy-tail distribution and then, $p = f(s,n)$, where $f(s,n) = 1$ when $n>0$, and otherwise it follows the procedure described in \citet{dabney2021ez_greedy}. In addition, ADEU can also improve $\epsilon z$-greedy by incorporating more complex definitions of uncertainty, maintaining the use of the options. Hence, ``extensions'' of $\epsilon z$-greedy \cite{biedenkapp2021temporl,Lee2024UTE}, that improve~\citet{dabney2021ez_greedy} by modifying the way that the length $n$ of a trajectory is decided, can also be incorporated by ADEU.
In addition, instead of having a Bernoulli distribution, we can utilize a Categorical one. The process of modifying $p$ will remain the same, however, in this scenario, we can simplify the ``task'' of finding an option. 

Given that ADEU can subsume existing adaptive exploration methods as special cases, it is reasonable to expect that it can replicate their performance in the tested domains. Thus, ADEU is well-suited for tasks with {\em both} dense and sparse reward structures. For instance, {\em ADEU %can 
solves
the extremely demanding DeepSea domain~\cite{osband2019deepSea} by emulating Dabney {\em et al.}'s~\cite{dabney2021ez_greedy} adaptive exploration method.} 

% In addition, ADEU can enhance exploration in such settings by using more complex definitions of ``uncertainty'' (e.g.,~\cite{nikolov2018informationdirected, Badia2020NeverGiveUp}). 

\paragraph{Motivating example cont'd:}\label{sec:motivatio_example_results}
%note: we have already said why adeu is ideal fir that example and where the other fail
We now continue with the technical setup and the results of the motivation example. We design a large-scale grid environment of size $1500 \times 1500$, extending the setup shown in Figure~\ref{fig:motivation_example}. In this environment, we evaluate four different agents: {\em (i)} a standard $Q$-learning agent using $\epsilon$-greedy exploration, {\em (ii)} an intrinsic motivation (IM) agent built on top of $Q$-learning, {\em (iii)} a UCB-based ensemble agent composed of multiple $Q$-learners, and {\em (iv)} an instance of ADEU. For the IM agent, we adopt ideas from \cite{tang2017exploration} without the need for approximation due to the discrete state space. The ADEU agent adjusts its action selection based on state visitation frequency, favoring exploration in less-visited states and exploiting learned values in frequently visited ones. It uses the same mechanism as the IM agent. 
%%%%%%%%%%%%%%%%%%%
%Only for the arXiv Version
%(See Appendix for more details.)
%%%%%%%%%%%%%%%%%%%%%%%%%%

Table~\ref{tab:example_results} shows the maximum reward and the mean reward during learning, computed as follows. We employed $5$ different runs (seeds) for all agents, with each run comprising $3\times10^4$ episodes. An episode terminates when: {\em (i)} the agent reaches the goal; or {\em (ii) } the agent falls in a hole; or {\em (iii)} the agent acts/survives $t>3\times1500$.
The following process was performed for each run: 
Once every $100$ episodes, an ``evaluation'' episode was executed, in which the agent greedily used its policy (i.e., no actions were selected for exploration during that episode); and we stored the reward accumulated during that episode, and all such episodes in each run.
We then created an ``average run'' by averaging over these episodes' values across all runs;
and computed the mean episodic reward and the maximum episodic reward in that average run. Note that the maximum episodic reward was achieved after convergence.

\begin{table}[h!]
\centering
\caption{\label{tab:example_results}Mean and Max episodic rewards 
in the Frozen Lake domain (average over $5$ runs). ADEU was built on top of a $Q$-learning agent.}
\resizebox{0.6\linewidth}{!}{%
\begin{tabular}{l@{\hspace{8mm}}l} 
\toprule
\textbf{Exploration Strategy} & \textbf{Mean and Max episodic rewards} \\
\midrule
$\epsilon$-greedy & 334; \,\,\,983 \\
\rowcolor{gray!10}
ADEU & 88,854; \, 104,487 \\
IM~\cite{tang2017exploration} & 393; \,\,\,989 \\
\rowcolor{gray!10}
UCB~\cite{Chen2017ucb} & 36,070; \, 104,487 \\
\bottomrule
\end{tabular}%
}
\end{table}

\begin{figure}[h!]
    \centering
    \includegraphics[width=0.5\linewidth]{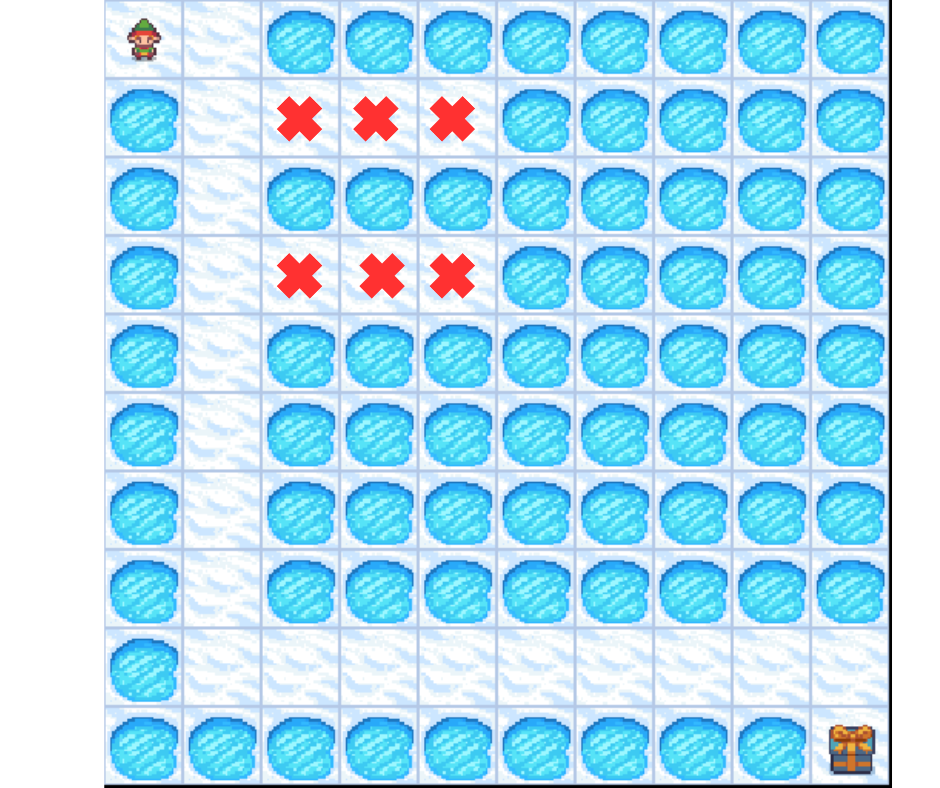}
    \caption{Environment used in testing safety. In this particular figure, the dimensions of the grid are $10\times 10$. The red crosses represent the unsafe states.}
    \label{fig:safety_environment}
\end{figure}

As shown in Table~\ref{tab:example_results}, ADEU demonstrates superior performance compared to the other methods by rapidly learning a policy that consistently guides it toward the target price. Specifically, the ADEU agent avoids excessive exploration of frequently visited states—those characterized by low uncertainty according to the employed function $f(s)$. Instead, it tends to replicate the learnt trajectory until completion before initiating further exploration; and manages to learn the correct trajectory. By contrast, both the $\epsilon$-greedy and IM algorithms fail to learn a successful trajectory to the goal, likely due to their persistent over-explorative behaviour. Finally, UCB learns the correct trajectory, but it requires nearly four times as many episodes as ADEU (16,800 vs 4,515 for ADEU). Its average reward is approximately half that of ADEU's. Note that in our main experiments involving more complex action and state spaces, UCB always ranks behind both ADEU variants (see Table~\ref{tab:results}).

\begin{figure}[h!]
    \centering    
    \includegraphics[width=1\linewidth,height=0.65\linewidth]{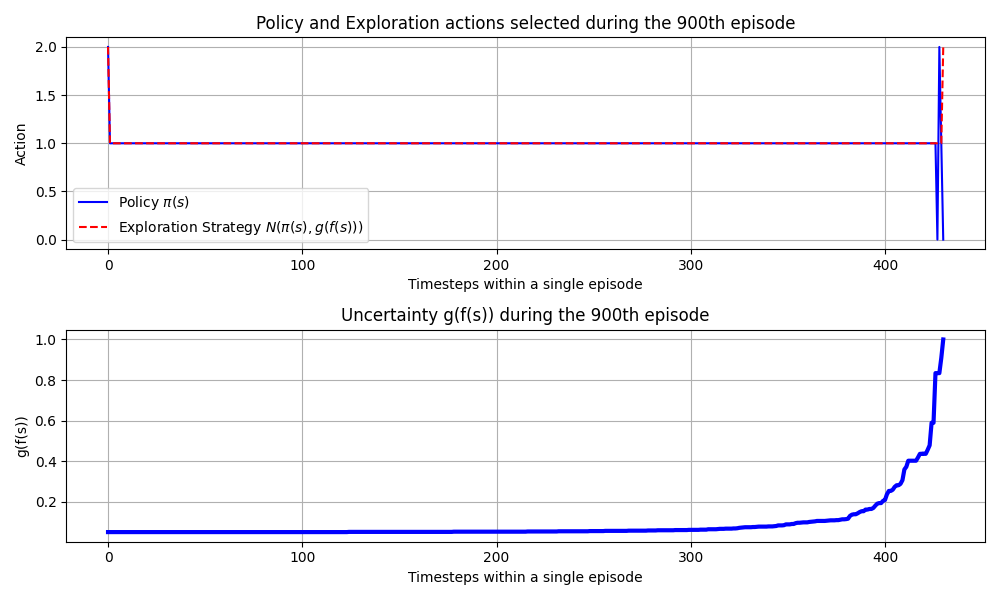}
    \caption{(Top) Blue line shows $\pi(s)$ across one episode. The red line shows the action selected by the agent. (Bottom) Uncertainty as calculated by the agent across the states of a single episode.}
    \label{fig:exploration_rate}
\end{figure}

Figure~\ref{fig:exploration_rate} (top subfigure) shows both the actual policy $\pi(s)$ and the action selected from the ADEU agent (so, using Equation~\ref{eq:adeu}) during a single episode in our Frozen Lake environment. In addition, the bottom subfigure shows the value of $g(f(s))$ during the same episode. 
As expected, at the beginning of the episode, $g(f(s))$ outputs low values since the agent visits states that it is certain about. Hence, the action selected by the ADEU agent coincides with %the one
that dictated by $\pi(s)$. As the agent visits states with high uncertainty, it does not trust its so far learned policy and starts exploring them. As of that, the action selected by ADEU and the policy $\pi(s)$ differ at the end of the episode, since $g(f(s))$ increases. 
%We should also note
Note also that the action suggested by $\pi(s)$ in the later timesteps of the episode is not correct. However, the action sampled for exploration was the correct one, allowing the agent to ``survive'' for more timesteps and {\em gain useful experience}. Thus, indeed, the exploration phase should have begun at the later timesteps; had it started earlier, the agent would not have ``survived'' to extend the known optimal policy.
% for the early timesteps to the later ones. 
\footnote{We expect the same behaviour in more difficult multi-dimensional domains such as the ones described in Section~\ref{sec:experiments}. However, we cannot create a figure similar to Figure~\ref{fig:exploration_rate} for those, since both $\pi(s), a \in  \R^n$, where $n>2$.}

\subsection{ADEU for safe exploration}

Now, as discussed and seen in the example, ADEU avoids exploration in states with low uncertainty. Moreover, it can be extended so that it conducts {\em safe exploration}---i.e., to not explore in {\em unsafe} states~\cite{wachi2023safe,ijcai2022p3o}.
%Intuitively, to keep safe, there exist states that the agent should not visit. 
By simply defining a new mechanism that also considers this extra constraint, ADEU can avoid exploring these states.

More precisely, suppose that a mentor has already informed the agent about some potential unsafe states (like in~\citet{thomas2021imagining_near_future}) {\em and} the policy $\pi(s)$ the agent should follow in these particular states. We would like the agent to {\em not} explore finding a better policy in these specific states. Hence, by designing an $f(s)$ that outputs low uncertainty in these states, an ADEU agent can alternate between exploration and exploitation proportionally to the safety of the current state. Such an ADEU agent, will simply follow Equation~\ref{eq:adeu}.

%Similarly
Alternatively, the agent can be informed about the cost function of the environment (like in~\citet{tian2024reinforcement}). By replacing $f(s)$ with the cost function, ADEU can combine adaptive exploration with safety. Even if the ADEU agent {\em is not informed} regarding the unsafe states or the cost function, it can interact with the environment to approximate it. 
Along with this ``approximation'' process, the agent can still respect the safety constraints. 
In that case, ADEU only requires an initial safe %(yet not necessarily optimal) 
policy as in \citet{as2025actsafe}. That is, Eq.~\ref{eq:adeu_safety} allows for the aforementioned
``safety-driven'' or ``safety-aware'' behaviour, without requiring knowledge of the cost function. 
\begin{equation}\label{eq:adeu_safety}
    a \sim \mathcal{D}\left (\pi(s), g(\nicefrac{1}{f_\text{safety}(s)})\right )
\end{equation}  
where $f_\text{safety}(s)$ is the expectation of the environment's cost function $ \E_{s,a}\max_a\{ g_\text{cost}(s,a)\}$ (where $g_\text{cost}(s,a)$ follows the standard definition described in~\cite{altman1999cmdp}), which is learned by interactions. 
%All other notation follows the one in Equation~\ref{eq:adeu}.

By sampling from Equation~\ref{eq:adeu_safety}, the agent will replicate the known policy $\pi(s)$ in a state with high expected cost; in any other state, the agent will explore randomly. 
In more detail, in a state with high $f_\text{safety}(s)$, the sampled action $a$ will not deviate from $\pi(s)$, given the low variance in Equation~\ref{eq:adeu_safety}. By contrast, in a state with low $f_\text{safety}(s)$, the sampled action will promote exploration.
Finally, Equation~\ref{eq:adeu_comb} shows how we can combine Equations~\ref{eq:adeu} and~\ref{eq:adeu_safety} to achieve {\em both} {(i)} {\em safe exploration} when required; and (ii) {\em uncertainty-driven adaptive exploration} in states where safety does not matter: 
\begin{equation}\label{eq:adeu_comb}
    a \sim \mathcal{D}\left (\pi(s), g(\nicefrac{ f_\text{uncertainty}(s)}{\lambda f_\text{safety}(s)})\right )
\end{equation}  
In Eq.~\ref{eq:adeu_comb} $f_\text{uncertainty}(s)$ is the aforementioned uncertainty-measuring mechanism and all other notation (except $\lambda$) follows Eq.~\ref{eq:adeu_safety}. Note that $\lambda$ generally acts as a term (or mechanism) that ensures the safety term ``dominates'' when safety is critical. For example, $\lambda$ can take ``conditioned'' values  for a state $s$ e.g., $\lambda = [ max_a\{Q(s,a) \}$ if $\E_{s,a}\max_a\{ g_\text{cost}(s,a)\} > max_a\{Q(s,a)\} $, otherwise 1 $]$. Therefore, the $f_\text{safety}(s)$ term will be amplified when the expected cost is high.

In this case, in states with high expected cost, the agent will act according to the known policy, in other states, it will act according to uncertainty-driven adaptive exploration. Thus, Equation~\ref{eq:adeu_comb} proposes a general agent that can take the best of both worlds.

\begin{table}[h!]
\centering
\caption{\label{tab:safety_results}Performance metrics of different exploration strategies in the Safety Frozen Lake domain. Results are averages over $5$ runs. ADEU was built on top of a $Q$-learning agent.}
\resizebox{0.4\textwidth}{!}{%
\begin{tabular}{l@{\hspace{3.5mm}}c@{\hspace{1.5mm}}c} 
\toprule
\textbf{Exploration Strategy} & \textbf{Mean \& Max Reward} & \textbf{Mean Constr.} \\
\midrule
$\epsilon$-greedy & 210; 739 & 0.34  \\
\rowcolor{gray!10}
IM~\cite{tang2017exploration} & 267 ; 780 & 0.55  \\
UCB~\cite{Chen2017ucb} & 12,605 ; 65,687 & 0.10 \\
\rowcolor{gray!10}
ADEU (using Equation~\ref{eq:adeu}; $f(s) = $Visitation freq.) & {\bf 59,865; 104,487} &  0.01 \\
ADEU (using Equation~\ref{eq:adeu_safety}, i.e., safety-aware) & 7;7 &  \textbf{0.00} \\
\rowcolor{gray!10}
ADEU (using Equation~\ref{eq:adeu_comb}; $f_\text{uncertainty} = $Visitation freq.) & 19,255; 85,087 & \textbf{0.00} \\
\bottomrule
\end{tabular}%
}
\end{table}

\begin{table*}[h!]
\centering
\caption{\label{tab:results}Mean and Max episodic rewards in various domains~\cite{towers2024gymnasium}. Results are averages over $15$ runs.}
\resizebox{0.65\textwidth}{!}{%
\begin{tabular}{@{}lcccccc@{}}
\toprule
\multirow{2}{*}{\textbf{Domain}} & 
\multicolumn{3}{c}{TD3 + ADEU} & 
\multirow{2}{*}{TD3+UCB} & 
\multirow{2}{*}{TD3+RND} & 
\multirow{2}{*}{TD3+Noisy Nets} \\
\cmidrule(lr){2-4}
& $f(s)=c$ & $f(s)=\text{RND}$ & $f(s)=\lambda Q_{\text{std}}(s,\pi(s))$ & & & \\
\midrule
Walker2d       & 564; 726       & \textbf{745; 947}     & 413; 500       & 355; 397     &1; 23        & 519; 657 \\
Hopper         & 376; 631       & \textbf{733; 1167}    & 293; 514       & 245; 422     & 10; 35       & 429; 766 \\
Swimmer        & 40; 46         & \textbf{52; 62}       & 36; 39         & 32; 36       & -3; -2         & 30; 35 \\
Ant            & 345; 980       & \textbf{1072; 1341}   & \textbf{1134; 1500}       & 585; 790     & -2658; 980   & 364; 981 \\
Humanoid       & 86; 132       & 313; 345              & \textbf{512; 658} & 89; 149   & 85; 132     & 75; 88 \\
Hum. Stand.   & 61511; 64490   & {\bf 73895; 79978} & \textbf{72978; 80139} & 54928; 56527 & 55211; 56057 & 62152; 63645 \\
\bottomrule
\end{tabular}%
}
\end{table*}

Figure~\ref{fig:safety_environment} shows a variant of the modified Frozen Lake that also incorporates some constraints. In more detail, the red crosses represent non-terminal states that the agent should not visit. In cases that the agent visits such states, a {\em safety constraint} is {\em violated}, and also the agent gets an immediate reward equal to $0$. Otherwise, it is rewarded according to the previous example. In addition, we conducted experiments in a $1500\times1500$ grid similarly to Section~\ref{sec:motivatio_example_results}.

To test both aforementioned scenarios, we create $3$ different instances of ADEU: {\em(i)} an agent that has been pretrained on a safe policy in states neighboring unsafe states, and explores using the visitation frequency definition of uncertainty (Equation~\ref{eq:adeu}); {\em (ii)} a safe ADEU agent that has been pretrained on the same safe policy, but explores according only to the cost function (Equation~\ref{eq:adeu_safety}), which is learned simultaneously with policy $\pi(s)$; and {\em (iii)} a safe ADEU agent built on top of the aforementioned one, which explores the environment according both to the learned cost function and the visitation frequency definition of uncertainty (Equation~\ref{eq:adeu_comb}).

Table~\ref{tab:safety_results} shows the mean and maximum episodic evaluation rewards throughout the training process. In addition, it shows the mean episodic constraint violations. All ADEU instances outperform their opponents since they achieve fewer constraint violations. Since all algorithms tested in a domain requiring efficient exploration, the ADEU instance that only uses the learned cost function (Equation~\ref {eq:adeu_safety}) did not yield a high reward due to the lack of (efficient) exploration in safe states. By contrast, the ADEU instance of Equation~\ref{eq:adeu}  effectively explores the environment and finds the final goal faster, but at the cost of some constraint violations---still violates {\em an order of magnitude fewer constraints} than its non-ADEU opponents. Finally, the ADEU instance that combines safety with uncertainty-driven adaptive exploration (Eq.~\ref{eq:adeu_comb}) yields a high reward, achieving $0$ constraint violations, utilizing the best of both worlds (uncertainty-driven adaptive exploration {\em and} safe exploration).

% These results provide interesting intuitions for ADEU potential for  %requiring efficient {\em and}
% safe exploration. As discussed in Section~\ref{sec:conclusions}, we are actively working towards {\em theoretical guarantees} 
% for ADEU in safety domains.

\section{Experimental Evaluation}\label{sec:experiments}

To experimentally test our framework in earnest, we create two main instances of ADEU, using as $f(s)$: {\em (i)} the intrinsic motivation mechanism proposed in~\citet{burda2018rnd}; and  {\em (ii)} the epistemic uncertainty measuring mechanism proposed in~\cite{Chen2017ucb}. These are the agents described in Table~\ref{tab:simple_adeu_instances}. Both agents were built on top of a {\em TD3}~\cite{fujimoto2018TD3} DRL algorithm's original implementation.\footnote{\url{https://github.com/sfujim/TD3}.} 
As such, we refer to them as:
{\em (i)} TD3+ADEU with $f(s) = RND$;\footnote{More precisely but less concisely, ``TD3+ADEU with $f(s) = r_\text{intrinsic}^{RND}$''.}
% Reward function choices and implementation details for all agents are %provided 
% in the Appendix.} 
and {\em (ii)} TD3+ADEU with $f(s)=\lambda Q_\text{std}\left(s, \pi(s)\right)$. In addition, we consider an instance of ADEU with an {\em uninformative} $f(s)$, and create a ``TD3 + ADEU with $f(s)=c$'' agent, where $c$ is a constant. Obviously, $f(s)=c$ is {\em not} an appropriate uncertainty measuring mechanism.
The agent with constant $f(s)$ will serve as baseline, since its distribution matches the one used in TD3~\cite{fujimoto2018TD3} and hence, they share the same exploration strategy. We use $g(f(s)) = c$ for {\em this} ADEU instance.

For the other ADEU instances, we use Eq.~\ref{eq:g_s},
with
$c=0.2$ as in TD3 implementation. 
%This is only for the arXiv version
%%%%%%%%%%%%%%%%%%%%%%%%%%%%%%%%%%%%%%
%(More details regarding the choice of $g(\cdot)$ are in the Appendix.)
%%%%%%%%%%%%%%%%%%%%%%%%%%%%%%%%%%%
\begin{equation}
\label{eq:g_s}
g(f(s)) = sigmoid(f(s))  \cdot c
\end{equation}

%We built all of our agents on top of a TD3 agent, implemented as in the TD3 github.\footnote{https://github.com/sfujim/TD3}. 
Our agents use TD3 as a ``basis'' algorithm.
In principle, any DRL algorithm could have been used instead. 
Since ours is an exploration framework, we used a DRL algorithm %that is 
not tied to any exploration strategy---and TD3 uses a `vanilla' one that can be readily replaced with ADEU.
By contrast, SAC~\cite{Haarnoja2018SAC}  uses the entropy framework, and hence,
%its exploration strategy is `tied' with the agent's NN. 
it is tied to that framework's particular exploration strategy.

We also implement three agents to act as our main competitors, to test ADEU against representatives of three standard families of exploration methods: IM exploration, epistemic uncertainty-driven exploration, and (parameter-space) noise exploration. Specifically, we test against {\em TD3+RND}, an agent that acts greedily in the environment with a reward function equal to $r_\text{extrinsic}(s) + r^\text{RND}_\text{intrinsic}(s)$  (cf. Equations~\ref{eq:im-original} and~\ref{eq:rnd}).
In addition, we compare against {\em TD3+UCB}, an agent that selects actions aiming to maximize Equation~\ref{eq:ucb}.

Finally, for interest, we also create an agent that uses Noisy Nets~\cite{fortunato2018noisy} to perturb the actor's network and achieve more efficient exploration. We compare against this  {\em TD3+Noisy Nets} agent, since Noisy Nets is a popular choice for exploring robotic domains. Note that Noisy Nets does not utilize any definition of uncertainty---it is just an ``improved'' TD3 agent in terms of exploration.

We 
first
conducted 
%our 
experiments in standard MuJoCo domains~\cite{towers2024gymnasium}. Specifically, we used the five domains considered the most difficult according to the official documentation.\footnote{\url{https://gymnasium.farama.org/environments/mujoco/} . We did not use~\cite{gymnasium_robotics2023github} since its domains do not have the ``early termination'' condition.} Such domains do not satisfy Definition~\ref{def:game}. However, the ADEU agents still achieve (as later described) superior performance against their opponents.
In these domains, the agent should explore the {\em continuous multidimensional action space} to find the best policy. A dense reward function is used to describe these robotic tasks. However, ADEU can also be used in environments with a sparse reward function, when using appropriate uncertainty-measuring mechanisms e.g.,~\cite{Badia2020NeverGiveUp, nikolov2018informationdirected}.

Table~\ref{tab:results} presents agents' performance in terms of mean and max episodic accumulated reward in the aforementioned domains. These mean and maximum values were calculated over $15$ runs (seeds), via the process described in the motivation example. We present both maximum and mean reward to track both the best performance of an algorithm and an indication of the time required to reach convergence.
We mark in bold the best-performing algorithms in terms of mean episodic reward within each domain---an algorithm is considered to be ``best-performing'' if its mean episodic reward reaches at least $90\%$ of the highest mean episodic reward accumulated in the domain. 
%Only for the arXiv Version
%%%%%%%%%%%%%%%%%%%%%%%%%%%%%%%
%{\em The (standard) figures where the cumulative episodic evaluation reward is plotted can be found in the Appendix.}
%%%%%%%%%%%%%%%%%%%%%%%%%%%%%%%

\paragraph{ADEU improves DRL performance:}
As seen in Table~\ref{tab:results}, {\em TD3+ADEU} with $f(s)=RND$ or {\em TD3+ADEU} $f(s) = \lambda Q_{std}(s, \pi(s))$ outperform the others in {\em every} domain. In addition, in the three most difficult domains (i.e., Ant, Humanoid Standup, and Humanoid) {\em both} of those ADEU instances outperform {\em all} other agents, achieving the first and the second best performance.  It is also noteworthy that an instance of ADEU
equipped with
a specific uncertainty-measuring mechanism $f(s)$ typically outperforms the original exploration method that employs the corresponding uncertainty-measuring mechanism.

In addition, our results show that ADEU %combined 
with  $\lambda Q_\text{std}(s, \pi(s))$ outperforms most of its opponents in the 
three 
aforementioned highly challenging 
%Ant, Humanoid, and Humanoid Standup 
environments. This indicates this mechanism tracks uncertainty more efficiently in larger environments.

\begin{figure}
    \centering
    \includegraphics[width=1\linewidth,height=0.5\linewidth]{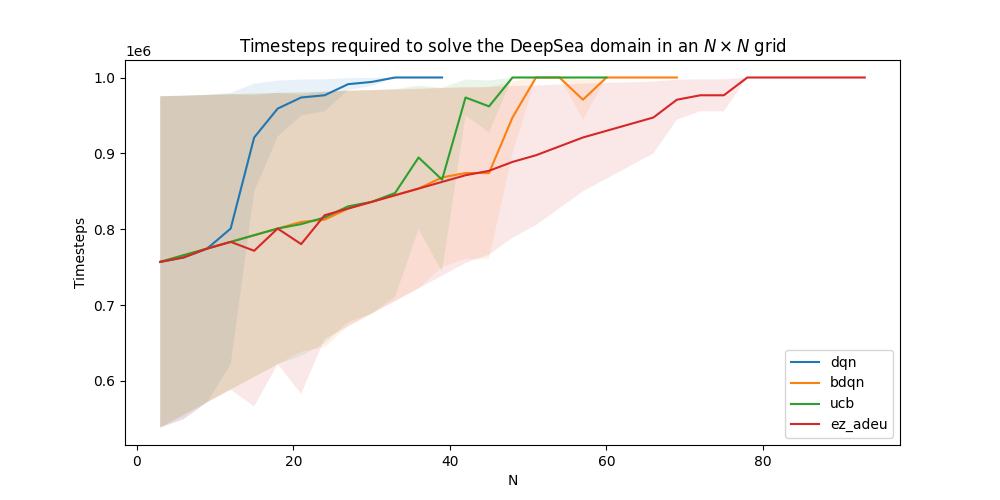}
    \caption{\label{fig:deepsea_results}Timesteps required for each algorithm to solve the DeepSea domain for $N\times N$ grid. {\em The lower the curve, the better the performance of the algorithm.}}
\end{figure}

\paragraph{ADEU is a `plug-and-play' adaptive exploration framework:}
As seen in our experiments, ADEU can incorporate any uncertainty-measuring mechanism of choice. Of course, as also discussed above, not all uncertainty-measuring mechanisms are created equal. Notice, e.g., that the {\em naive} instance of {\em TD3+ADEU with $f(s)=c$} is outperformed by at least one other ADEU instance in each domain. This indicates that ADEU efficiency depends on the efficiency of the uncertainty mechanism. Seen otherwise, the choice of an appropriate uncertainty-measuring mechanism for use in ADEU, promises the improvement of performance in the domain of interest. 
\paragraph{Results in the DeepSea domain:} 
%In this experiment, 
We also employed ADEU in DeepSea,
%a domain which 
in order to
put the agent's ability for deep exploration to 
an arduous test. 
Specifially, we tested the emulation of~\citet{dabney2021ez_greedy} from the ADEU framework, hereafter referred to as $\epsilon$z-adeu, in this demanding domain. 
%%%%%%%%
%This is only for the arXiv version
%For theoretical results and details of the implementation, please refer to the Appendix. 
%%%%
Figure~\ref{fig:deepsea_results} shows the timesteps required for an algorithm to solve (i.e., find the tressure) a DeepSea domain with a $N\times N$ grid. The $x$-axis of that figure shows the different values of $N$ while the $y$-axis shows the aforementioned timesteps. 

% We arbitrarily set the maximum allowed timesteps to $10^6$ in which the agent should reach the treasure for $10^5$ {\em consecutive} evaluation timesteps ($= 10\%$ of the training time). In any other case, the episode is considered as ``fail''. {\em That is, in such graphs, a lower curve denotes better performance for the agent}.

As seen in Figure~\ref{fig:deepsea_results}, $\epsilon$z-adeu achieves better performance in terms of scalability in the DeepSea domain. Specifically, $\epsilon$z-adeu {\em solves} the DeepSea domain for $N\approx 80$ when its best opponents struggle even for $N > 60$. We should also highlight that both the BDQN and UCB opponents were built on top of an ensemble of DQNs---an architecture that %achieves 
promotes
Deep Exploration---{\em in contrast} to $\epsilon$z-adeu, which was built on top of a DQN. 

%To conclude the experimental section, 
It is clear from our experiments that ADEU, %framework, 
paired with efficient uncertainty-measuring mechanisms, can be utilized for both demanding control tasks and domains requiring deep exploration.

\section{Related Work}\label{sec:related_work}

We now briefly discuss related work regarding IM, epistemic uncertainty, or adaptive exploration.

{\em Intrinsic Motivation:} 
%As discussed, most IM methods focus on discrete action space domains with sparse reward functions. However, some others works 
Several IM approaches have been tested in robotic domains. For instance,~\cite{Lobel2023FlippingCoins} %employs a Coin Flip Network to 
design an efficient intrinsic motivation function that can be utilized as an uncertainty mechanism in ADEU. They test their method in a sparse reward variance of the Fetch domain; an easier domain than standard robotics.
Then,~\cite{mahankali2024random} adds a random reward to the sparse reward function,
to guide the agent to explore different parts of the environment. This can be used 
by
ADEU  
in extremely sparse reward function domains.

{\em Epistemic Uncertainty}:  As mentioned earlier, UCB~\cite{Chen2017ucb,lee2021sunrise} is commonly used 
in continuous action-space robotic domains. We also created an instance of ADEU using this ensemble mechanism to track uncertainty (the ADEU instance with $f(s) = \lambda Q_\text{std}(s, \pi(s))$ in Sec.~\ref{sec:experiments}) and test it against a UCB implementation built on top of a TD3 agent. \citet{da2020uncertainty} ``asks'' a mentor for guidance when the uncertainty is high; uncertainty was measured in a similar way.  

{\em Active Exploration}: Several works~\cite{russo2023active1,hazan2019active2,hong2020active4,nikolov2018informationdirected} have proposed methods to efficiently explore the environment to derive an $\epsilon$-optimal policy, while also obtaining a low sample complexity.  Some methods~\cite{russo2023active1,hazan2019active2}, where tested in tabular or ``discretized'' environments. Hence, these works tackle a distinct exploration problem, and there is no straightforward comparison with ADEU. Other works~\cite{hong2020active4,nikolov2018informationdirected} incorporate these ideas in the classic DRL testbeds either by proposing mechanisms to motivate efficient exploration~\cite{nikolov2018informationdirected} or by changing the reward function that the agent aims to maximize~\cite{hong2020active4}. Both mechanisms can be used in the ADEU framework. 

{\em Adaptive exploration:}~\citet{tokic2010adaptive} uses the TD-Error to change the $\epsilon$ 
%parameter
of the $\epsilon$-greedy in each state. Hence, an agent is more likely to explore when uncertainty about the $Q$-function is high. %This method was tested in a $10$-armed bandit task (with discrete state and action spaces). 
\citet{pislar2022when} is the first work that actively attempts to create a 
%modern 
framework for {\em intra-episodic} exploration. 
They propose numerous external ``signals'' for changing modes from exploitation to exploration, either heuristic or random ones. As noted earlier, these can be used as ADEU uncertainty mechanisms. Similarly to~\cite{pislar2022when}, $\epsilon$z-greedy~\cite{dabney2021ez_greedy} suggests following directed behaviour with random---sampled from a heavy-tailed distribution---length. This directed behaviour guides the agent to repeat a single action numerous times. %Thus, it achieved excellent results in settings which require replicating the same action numerous times, such as  DeepSea~\cite{osband2019deepSea} and  Chain Walk~\cite{osband2016bdqn}. %On a similar note,~\cite{lixandru2024ppo_adaptive} proposes an adaptive exploration variant of PPO~\cite{schulman2017ppo}, where the entropy term in the loss function is scaled by a heuristic based on the agent’s recently obtained rewards. Even though ADEU does not directly manipulate the loss, it can use this heuristic as $f(s)$.
Finally, most goal-based exploration works~\cite{ecoffet2019goexplore,Ecoffet2021first_return_then_explore,torne2023breadcrumbs} can be %considered 
viewed
as `adaptive exploration' ones since they replicate the known trajectory with no exploration, and then use some exploration to extend it.

\section{Conclusions and Future Work}\label{sec:conclusions}

In this paper, we put forward a generic adaptive exploration framework that uses a generic uncertainty-based action selection mechanism to decide in a principled manner when to alternate between exploration and exploitation.
The generality of that mechanism allows {\em (i)} our framework to effectively incorporate any uncertainty-measuring mechanism of choice; and 
{\em (ii)} existing adaptive exploration techniques to be viewed as special cases of our framework. 
In problems requiring adaptive exploration, ADEU is a promising and easily deployable solution. Additionally, it allows the user to select an existing uncertainty measurement mechanism or to define a heuristic one tailored to its problem. Our experiments verified the superiority of ADEU in difficult testbeds.

Ongoing and future work 
includes extending ADEU in various directions. To begin, we intend to perform tests to further verify ADEU's ability to exploit background knowledge; and its ability to recover from being fed with incorrect background information or to detect and escape clear suboptimal trajectories by optimizing the rollout procedure. %In addition, to make the learning procedure more robust, we will modify the learning process by weighting the learned experience before backpropagating it in the network, similar to~\cite{lee2021sunrise}. As mentioned earlier, we did not modify the learning procedure in this work, since ADEU does not exhibit instabilities, and we also aim to keep this work ``exploration-only''.
Moreover, we are actively working to extend ADEU 
to {\em guarantee}  safe exploration~\cite{wachi2023safe,ijcai2022p3o}.
In ongoing work, we are calculating the upper bound of allowed exploration in neighboring safe states, so that ADEU can guarantee safe exploration. We also aim to extend ADEU to multiagent environments. By leveraging agents' independent exploration, ADEU agents can adaptively explore the environment. Furthermore, by allowing a centralized agent to define $f_i(s)$ for each agent $i$, we can reduce the non-stationarity issues that usually arise in multiagent RL. Finally, we intend to design mechanisms to appropriately set the third and fourth moments of $D$, to {\em guide} exploration to particular directions.

\begin{acks}
%\section*{Acknowledgments}
The research described in this paper was carried out within the framework of
the National Recovery and Resilience Plan Greece 2.0, funded by the European Union - NextGenerationEU (Implementation Body: HFRI. Project name:
DEEP-REBAYES. HFRI Project Number 15430).
\end{acks}
% \begin{acks}
% If you wish to include any acknowledgments in your paper (e.g., to 
% people or funding agencies), please do so using the `\texttt{acks}' 
% environment. Note that the text of your acknowledgments will be omitted
% if you compile your document with the `\texttt{anonymous}' option.
% \end{acks}

%%%%%%%%%%%%%%%%%%%%%%%%%%%%%%%%%%%%%%%%%%%%%%%%%%%%%%%%%%%%%%%%%%%%%%%%

%%% The next two lines define, first, the bibliography style to be 
%%% applied, and, second, the bibliography file to be used.

\bibliographystyle{ACM-Reference-Format} 
\bibliography{sample}

\clearpage
\appendix
% After \appendix, before definitions:
\setcounter{definition}{0}     
\setcounter{corollary}{0}

\renewcommand{\thecorollary}{\arabic{corollary}}% optional: restart numbering
\renewcommand{\thedefinition}{\arabic{definition}} % just 1, 2, 3...

\section{Technical details regarding the ``modified Frozen Lake'' motivating example}\label{sec_app:motivation_example}

\begin{figure}[h!]
    \centering
    \includegraphics[width=0.6\linewidth]{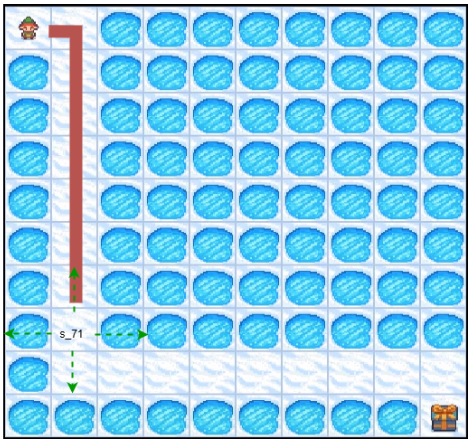}
    \caption{Environment used in our motivation example. In this particular figure, the dimensions of the grid are $10 \times10$. In our example, we use a $1500\times1500$ grid.}
    \label{fig:mot_example}
\end{figure}

For our motivation example, we create an environment similar to Figure~\ref{fig:mot_example}, but we increase the dimensions of the grid to  $1500\times1500$. As mentioned in the main text, we test $4$ different agents in the {\em modified} (as described in the main paper) Frozen Lake environment:\footnote{We note that all agents were also tested in the {\em classic} Frozen Lake, for fine-tuning debugging reasons and for fine-tuning their parameters. Hence, we do not present these results here.} {\em (i)} the classic Q-learning agent paired with $\epsilon$-greedy; {\em (ii)} an IM agent built on top of the Q-learning agent; {\em (iii)} a UCB agent consisting of $5$ Q-learning agents; and {\em (iv)} an ADEU agent built on top of the Q-learning agent, with an $f(s)$ as described below.

%reward function and other details

As discussed in the main text, in our motivation example, we used a dense reward function that rewards the agent when it reaches a state closer to the goal. In more detail, we reward the agent every time it performs a `forward' step and reaches a valid, {\em non-terminal} state $s$ according to Equation~\ref{eq:reward_frozen_lake}:

\begin{equation}
    \label{eq:reward_frozen_lake}
    r(s) = 1500 \cdot \frac{s}{\sum_{i=0}^{i = 1500^2} i}
\end{equation}
where $s$ is the id of the current state (i.e., a number between $0$ and $1500^2 - 1$); and $\sum_{i=0}^{i = 1500^2} i$ serves as a normalizer.

In the case of a ``hole'', the agent receives a penalty $(=-10)$, and the episode is terminated; while when the agent reaches the {\em goal} terminal state, it receives a high reward---equal to $10^4$ in our case. 
%Equation~\ref{eq:reward_frozen_lake} shows the reward %yielded 
%received
%by the agent when it reaches a non-terminal state.

As an intrinsic reward mechanism for the IM agent, we use the visitation frequency, calculated as $\frac{\beta}{ \sqrt{n(s)}}$, where is the exact visitation count $n(s)$. This approach is similar to~\cite{tang2017exploration}, but without the need to approximate the visitation frequency, as  in this tabular scenario we can in fact keep track of the times that an agent has visited a state $s$. 
As such, the reward function used for the IM agent is :
\begin{equation}
    r(s) = r_\text{extrinsic}(s) + r_\text{intrinsic}(s) = r_\text{extrinsic}(s) + \frac{\beta}{\sqrt{n(s)}} 
\end{equation}
where $\beta$ is a constant number between $0$ and $1$, and $n(s)$ is the aforementioned visitation frequency.

For the ADEU agent used in the motivation example, we select as $D$ a {\em categorical distribution} with $K$ categories, for which the probability of sampling-out an action $a$ (out of $K=4$ possible actions) at a specific 
state $s$ is determined by:

%uses a state-specific $p(s)$ for determining the probabilities by which we sample-out actions at each particular state $s$. In more detail, actions are sampled according to:

\begin{equation}
    p(s,a)   = sigmoid \left (\beta \sqrt{n(s)} \right ) \cdot q(s,a)
\end{equation}
where $q(s,a)$ is %a vector with 
the $Q$-values for a state $s$, $\beta$ is a constant number between $0$ and $1$, and $n(s)$ is the aforementioned visitation count.
We note that $p(s,a)$ will be normalized before action selection.
Note that, for the categorical distribution, 
if the agent's uncertainty is high at $s$, all $p(s,a)$ should have similar values, close to $1/K$; while if the agent's uncertainty is low at $s$ the $a$ with the highest $p(s,a)$ will be regularly selected.\footnote{The variance of observing the $X_i$ random variable corresponding to selecting $a$ is quite small in this case---specifically, equal to $p(s,a) (1- p(s,a))$.}
%In addition, we clarify that ${\bf p}(s)$ is a {\em vector} of probabilities where $p_i(s)$ is the probability of selecting action $i$ in state $s$. 

%As such, if $q(s,a)$ has positive values, $g(f(s))$ is bounded between the lower bound of the sigmoid function for positive input, and $q(s,a)$. 

Now, if the agent has visited a state $s$ numerous times, and hence $n(s) \rightarrow \infty$, it will select actions with probability $p(s,a) =  q(s,a)$. This is almost equivalent to the Boltzmann exploration with $T=1$. 
Consider now the case $n(s)=0$.
The ADEU agent should uniformly sample an action in this case, to achieve full exploration. 
However, if $n(s)=0$, then if the {\em classic} sigmoid function is used, it holds that 
$ p(s,a)$ is bounded by $0.5 q(s,a)$ (since $0.5$ is the lower bound of the classic sigmoid for positive input). %That means that the agent will sample actions without completely trusting the $Q$-values for that state. 

To mitigate this issue, we employ a different sigmoid function (we henceforth refer to this as ``shifted sigmoid'') that shifts the input before applying the classic sigmoid. 
We selected this ``shifted'' sigmoid since it has a lower than $0.5$ lower bound even when $x$ is positive. This will result to pre-normalized $p(s,a)$ values that are close to $0$ when $n(s)=0$; and, thus, following normalization to $p(s,a)$ values that are close to $1/K$ each for a particular $s$. Hence, it allows for more exploration when the visitation count of a state is low.

That is, for our ADEU agent in this motivation example, we sample actions at each state according to:
\begin{equation}\label{eq:motivation_example_adeu}
     p(s,a) = shifted\_sigmoid \left( \beta \sqrt{n(s)}\right ) \cdot  q(s,a) 
\end{equation}

\begin{figure}[h!]
    \centering
    \includegraphics[width=1\linewidth]{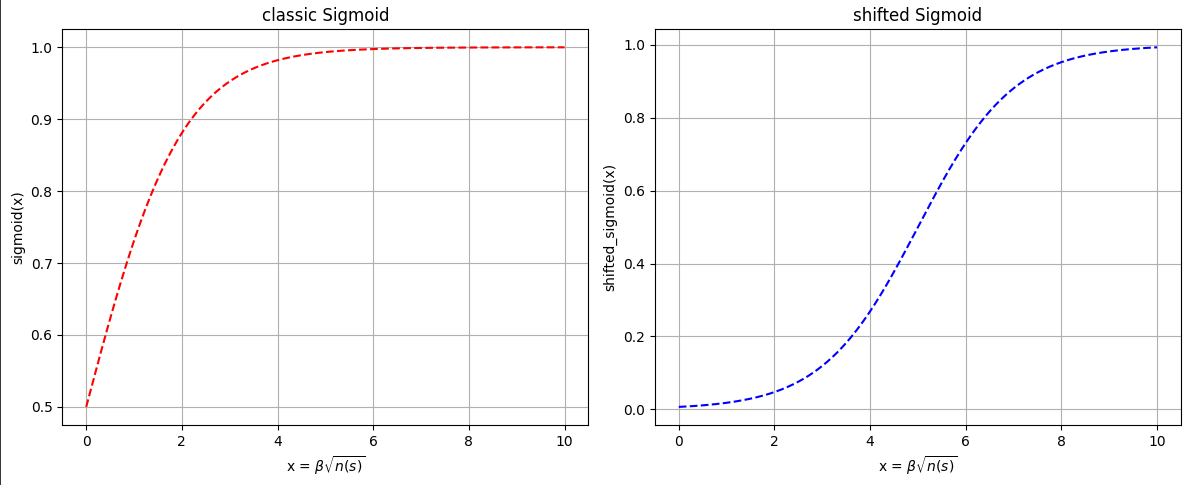}
    \caption{The ``shifted'' sigmoid function compared to the classic sigmoid.}
    \label{fig:sigmoid}
\end{figure}

Figure~\ref{fig:sigmoid} shows the two sigmoid functions. 

% This is so that, in the case of the categorical distribution, $g(\cdot)$ 
% %(i.e., to 
% %can be used for the categorical distribution case), the $g(x;s)$ 
% outputs a vector corresponding to the probability (after some normalization) that a specific action is sampled from the distribution given a state $s$ and $x$ input pair.

% The ADEU instance used in the motivation example, can also be seen as Boltzmann exploration where the temperature parameter is modified during the episode. Early in training, when the agent has visited a state only a few times, its confidence in the corresponding $Q$-values is low. To reflect this uncertainty, it scales the $Q$-values by a factor less than $1$, effectively reducing the differences between them and promoting uniform sampling. As the agent gains more experience and its confidence increases, this scaling factor approaches 1, allowing the agent to rely more heavily on the estimated $Q$-values when making decisions.

\section{Ablation Study on the Normalizer Function}~\label{sec:ablation_g}
%As previously mentioned, 
Eq.~\ref{eq:g_s} shows the $g(\cdot)$ used in our main experiments. 
\begin{equation}\label{eq:g_s}
g(f(s)) = sigmoid(f(s))  \cdot c
\end{equation}
%where $sigmoid$ is the standard sigmoid function and $c=0.2$. 
Since $c = 0.2$, $g(x)$ is bounded between $[0.1, 0.2]$, given that $x = f(s)$ is always positive. 
%With the same motivation as in the previous example, 
We also explored the possibility of using a custom sigmoid function, $\hat{g}(x)$:
\begin{equation}\label{eq:custom_g}
    \hat{g}(x) = (1 - \exp{(-x)})\cdot c
\end{equation}
where $x=f(s)$ and again $c=0.2$.

\begin{figure}[h!]
    \centering
    \includegraphics[width=0.8\linewidth]{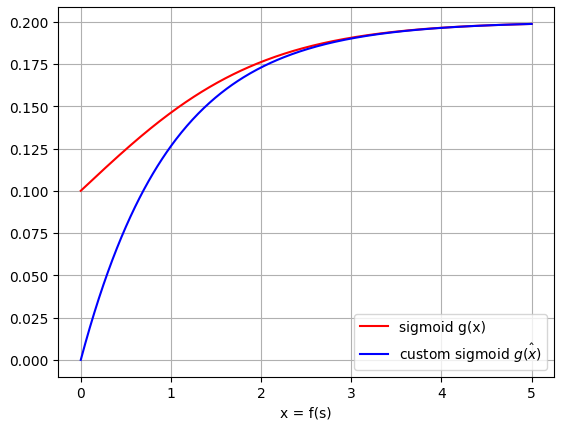}
    \caption{The two sigmoids tested in our {\em main} experiments}
    \label{fig:main_exp_sigmoids}
\end{figure}

Figure~\ref{fig:main_exp_sigmoids} shows the (positive input part of the) aforementioned functions: 
%In more detail,
the custom sigmoid $\hat{g}(x)$ will output a $0$ value in case of a low positive input value---i.e., when that agent has no uncertainty ($f(s)=0$), the variance of the distribution would be set to $0$. 

While this approach has desirable properties, it may lead to unintended consequences. Specifically, when uncertainty is low, the agent may repeatedly generate identical or highly similar experiences. This lack of diversity in the training data can increase the risk of catastrophic forgetting, since the agent produces correlated experience~\cite{Bruin2015catastrophic_forgetting_2}---at least for the trusted policy. 

 {\em This is why in our experiments we resolved to {\bf not} using the custom sigmoid function, but use the function of Equation~\ref{eq:g_s} as our $g(\cdot)$, as previously indicated.}
 Since the primary objective of our work in this paper is to address in a principled manner the problem of adaptive exploration, we were content with selecting a suitable form for $g(\cdot)$, i.e., that of Equation~\ref{eq:g_s}; and did not attempt to move the focus away from exploration itself in order to optimize performance.
Regardless, in future work, we plan to re-visit the $\hat{g}$ approach and attempt to avoid the catastrophic forgetting problem as follows.

A first approach is to attempt to modify the mechanism responsible for inserting ``known'' experiences into the replay buffer, by inserting an experience with a probability proportional to $\hat{g}(\cdot)$
(i.e., experiences relating to high uncertainty will be inserted into the buffer with high probability). A second way would be to attempt to adjust the loss function to mitigate overconfidence. 
This will implicitly modify the learning rate for the ``known'' experience, avoiding catastrophic forgetting, as in~\cite{kirkpatrick2017overcoming_catastrophic_forgetting}.

\section{Implementation details on RND's networks}

As discussed in the main text, the novelty detection mechanism used in RND for providing the intrinsic reward consists of two networks: {\em (i)} the Target; and {\em (ii)} the Predictor. Both networks use a state $s$ as input. The output of the Target network is a constant but randomly selected value, while the Predictor aims to predict it. Eventually, for the non-novel states, Predictor will converge to the Target's output. The convergence speed is highly affected by the hyperparametrs of the two networks. Instead of modifying these numerous parameters (e.g., learning rate of the Predictor, number of layers of each network, type of the optimizer etc) we have introduced a parameter that scales the output of both networks, termed $U$. That is, the final uncertainty-measuring mechanism used in the corresponding ADEU instance is the following:
\begin{equation}\label{eq:our_rnd}
    f_\text{RND}(s) =r^\text{RND}_\text{initrinsic} =  \left\lVert U\cdot f_{\text{predictor}}(s) - U\cdot f_{\text{target}}(s) \right\rVert_2^2
\end{equation}
We observe the behavior of the ADEU instance for numerous values of $U$, and we selected $U = 187.5$. This value was set for {\em all} the experiments.\footnote{In addition, we create a mechanism able to dynamically fine-tune $U$ throughout the training process by decreasing it if $f_\text{RND}(s)$ is high and increasing it if it is too low. We observed that this heuristic modification was used by ADEU only in the later stages of training (approximately after $85\%$ of the training).} We note that this modification (employing $U$) was {\em not} used in the TD3+RND agent since this would have affected the robustness of the agent.

\section{Proof of Theorem 1}

\begin{proof}
    To prove Theorem~\ref{theorem}, we will use mathematical induction. 
    \begin{itemize}
        \item {\bf Base Case ($i=0$):} The agent will start executing action $b_0$, after (randomly) exploring $s_0$ for a finite amount of time. 
        In addition, it will reduce its exploration of new actions at $s(0)$ since $f(s_0)$ decreases, thus learning $\pi(s_0) = b$. 
        (The frequency by which the execution of $b$ will increase and the exploration of other actions at $s$ will decrease, is relative to the effectiveness of $f(s)$).
        Moreover, the agent will start exploring actions at $s^*_1$ (due to transitioning there via $b$).
        \item {\bf Inductive Hypothesis ($i=k$):} We assume that $f(s_j) \rightarrow 0 \forall j=0...k-1$, and learned $\pi(s_j^*) = b_j$; and that the agent has found the action $b_k$ at $s_k$. As such, it decreases exploration in state $s_k$.  
        The agent will reach $s_{k+1}^*$ for the first time (due to transitioning there via $b_k$). 
        \item  {\bf Inductive Step ($i = k + 1$):} 
        % Starting from $s_0$ the agent will replicate $\pi(s_0), \text{ }\pi(s_j^*) \text{ } \forall j = 1...k$ since $f(s_0), \text{ } f(s_j^*) %\rightarrow 0$, and 
        Due to the inductive hypothesis, the agent will reach $s_{k+1}^*$ for the first time. Then, since $f(s_{k+1}^*) >> 0$ (there is high uncertainty in $s_{k+1}^*$, the agent will explore state $s_{k+1}^*$ until eventually finding $b_{k+1}$ and eventually learning $\pi(s_{k+1}^*) = b_{k+1}$. In addition, $f(s_{k+1}^*)$ will drop until reaching  $f(s_{k+1}^*) \rightarrow 0$, which will  prompt the agent to reduce exploration in $s_{k+1}^*$ and start exploring $s_{k+2}^*$. 
    \end{itemize}
\end{proof}

\section{Generalizing Definition 2.1:}

Definition 2.1, can be further generalized to also include sparse (but still increasing) reward settings. To show that, we will first define two types of states: 

\begin{definition}[Non-Terminal-Hyper-State]\label{def:non_terminal_hyperstate}
    A Non-Terminal-Hyper-State is defined as a set consisting of: {\em (i)} $N-1$ states $s_n : r(s_n, a) = 0 \forall a $; and {\em (ii)} one state $s_d : \exists a_d \text{ where } r(s_d, a_d \neq 0)$. All $s_n \text { and } s_d$ are {\em non}-terminal states.
\end{definition}

\begin{definition}[Terminal-Hyper-States]\label{def:terminal_hyperstates}
    A Terminal-Hyper-State is defined as the set of consecutive terminal states.
\end{definition}

Using the aforementioned definitions of Hyper-States, Definition~\ref{def:game} can be generalized also to include ``Sparse Increasing Reward functions Single Agent Games'' as follows:

\begin{definition}[Sparse Increasing Reward Functions Single Agent games]\label{def:sparse_reward_game}
    We define as ``Sparse Increasing Reward Function Single Agent Game'' a game that replaces states with Hyper-States in Definition~\ref{def:game}. In more detail, $s^*$ states are replaced by Nom-Terminal-Hyper-State with $r(s_d, a_d) > 0$, $s^!$ by Non-Terminal-Hyper-State with $r(s_d, a_d) < 0$ and all the other states are replaced by Terminal-Hyper-States.
\end{definition}

\begin{corollary}
    Theorem~\ref{theorem} holds for finding the best hyper-action in Sparse Increasing Reward Functions Single Agent Games. A hyper-action can be defined as a sequence of actions, exactly like ``options''. The uncertainty-measuring mechanism in ADEU variances that act in games satisfying Definition~\ref{def:sparse_reward_game} should calculate uncertainty in Hyper-States (i.e., a batch of non-terminal states). 
\end{corollary}

\section{Hyperparameters and Hardware details}

Table~\ref{app_tab:drl_params} lists all hyperparameters used in our experiments. 
To fine-tune the hyperparameter $\lambda$ for both TD3+UCB and ADEU with $f(s) = \lambda Q_\text{std}(s, \pi(s))$, we conducted experiments in the Hopper domain. We selected $\lambda = 10$ for TD3+UCB, as this value led to the best performance for TD3+UCB in this domain. For ADEU, we chose $\lambda = 1$ for the same reason. We note however that even with $\lambda = 10$, ADEU with $f(s) = \lambda Q_\text{std}(s, \pi(s))$ consistently outperformed TD3+UCB (with $\lambda = 10$) in this environment.

Furthermore, we compared ADEU using $f(s)=\lambda Q_\text{std}(s,\pi(s))$ and $f(s) = f_\text{RND}$ in the Hopper domain. Based on our results in this domain, we set $\rho = 0.1$ for the former, and $\rho = 0.3$ for the latter. We also observe improved behaviour in experiments in more difficult domains (such as the Humanoid etc) when $\rho = 0$ and $f(s)=\lambda Q_\text{std}(s,\pi(s))$. We did not conduct experiments with $\rho = 0$ and $f(s) = f_\text{RND}$ in the difficult domains. However, we {\em only} report the results for $\rho = 0.1$ in the case of $f(s)=\lambda Q_\text{std}(s,\pi(s))$. In future work, we plan to further investigate the behaviour of the agent for different values of $\rho$.

\begin{table}[h!]
    \centering
    \caption{Parameters used in our experiments}
    \begin{tabular}{ll}
        \toprule
        \textbf{Hyperparameter} & \textbf{Value} \\
        \midrule
        Discount ($\gamma$)         & 0.99 \\
        Soft update rate ($\tau$)   & 0.005 \\
        Policy noise ($c$)          & 0.2 \\
        Noise clip                  & 0.5 \\
        Policy update frequency     & 2 \\
        Learning rate               & 0.00005 \\
        Batch size                  & 256 \\
        Buffer size                 & $10^5$ \\
        $U$                         & 187.5 \\
        $\lambda$                   & [1, 10] \\
        $\rho$                      & [0.1, 0.3] \\
        $\beta$ (for UCB's binomial) & 1\\
        \bottomrule
    \end{tabular}
    \label{app_tab:drl_params}
\end{table}

All of our code was implemented in Python using the $3.11$ version. In addition, we use the $2.3.1$ version of pyTorch~\cite{paszke2019pytorch} and the $1.26.1$ version of numpy~\cite{harris2020numpy}. We also use the $1.1$ version of gymnasium~\cite{towers2024gymnasium}, and we ran our experiments using the latest ($v5$) version of the domains. 
\newpage
\section{Extended results}
\begin{figure*}[h!]
    \centering
    \includegraphics[width=1\linewidth]{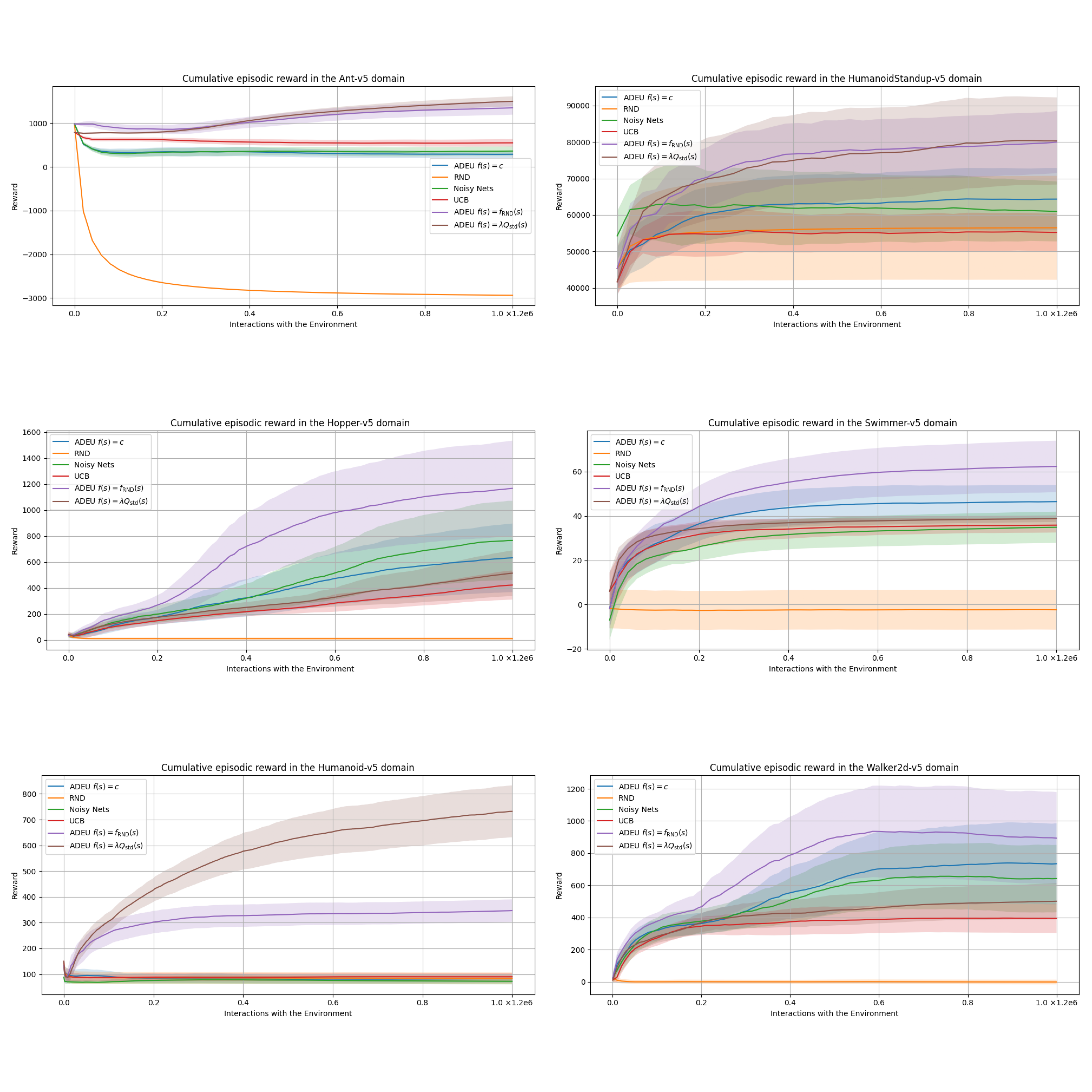}
    \caption{The cumulative episodic rewards yielded by each exploration strategy/framework in the tested domains. Results are averaged over $15$ seeds. Shaded areas correspond to the standard deviation.}
    \label{fig:extended_results}
\end{figure*}

Figure~\ref{fig:extended_results} shows the performance of each exploration strategy/framework in the tested domains. These results were used to calculate Table 4 of the main text. 

\section{Recreating ez-greedy as an ADEU instance}

Similarly to~\cite{dabney2021ez_greedy}, in $\epsilon$z-adeu there are $2$ distinct ``options'': {\em (i}) selection the ``go right'' action for $n$ consecutive timesteps; and {\em (ii)} select the ``go left'' option for $n$ consecutive timesteps. In both cases, $n$ is sampled from a heavy tail distribution (see~\cite{dabney2021ez_greedy} for more details). Regarding the DeepSea domain, we utilize the exact same environment described in~\cite{dabney2021ez_greedy}. As already discussed, a Distribution $D$, a policy $\pi(s)$, and an uncertainty-measuring mechanism $f$ are required, to define any instance of an ADEU agent. To emulate~\citet{dabney2021ez_greedy}, we will define the following parameters:

\begin{table*}[h!]
\centering
\caption{\label{tab:adeu_as_ez-greedy}
Emulating $\epsilon z$-greedy}
\renewcommand{\arraystretch}{1}
\begin{tabular}{|p{5cm}|p{5cm}|}
\hline
\textbf{Parameter} & \textbf{Selection} \\
\hline
Distribution $D$ & Bernoulli distribution. \\
\hline
Policy $\pi(s)$ & Equation~\ref{app:p_s_adeu_ez} \\
\hline
Uncertainty-Measuring Mechanism & Algorithm~\ref{alg:adeu_ez} \\
\hline
\end{tabular}
\end{table*}

Equation~\ref{app:p_s_adeu_ez} describes the stochastic policy of the ADEU agent that emulates the $\epsilon z$-greedy algorithm. In more detail, the agent utilizes a Bernoulli distribution that selects the ``exploitation'' action with probability $1-f(s,n,\cdot) $, or follows the given option for $n$ timesteps. Intuitively, ADEU selects an action that promotes exploration when the uncertainty $f(s,n,\cdot)$ is high.

\begin{equation}\label{app:p_s_adeu_ez}
     \pi(s) = \left\{\begin{matrix}
    \arg\max_i Q(s,a_i),\quad 1-f(s,n,\cdot) \\ \omega \sim U(A),\quad f(s,n,\cdot) 
\end{matrix}\right.
\end{equation}

where $f(s,n,\cdot)$ is the uncertainty-measuring mechanism described in Algorithm~\ref{alg:adeu_ez}, $Q(s,a_i)$ is the discrete action space Q-Function for each action $i$ derived from any DRL algorithm of choice, and $U(A)$ is the mechanism that calculates the options as described in~\cite{dabney2021ez_greedy}.

\begin{algorithm}
\caption{\label{alg:adeu_ez}Uncertainty measuring mechanism $f()$ for emulating $\epsilon$z-greedy~\cite{dabney2021ez_greedy}}
\begin{algorithmic}[1]
\State \textbf{Inputs:} $\epsilon$ (float): probability to begin an option (see~\cite{dabney2021ez_greedy}), $n$ (integer $\geq 0$): steps exploring using an option, \textit{explored} (boolean): indicates whether the agent explored or not during the previous step.

\State \textbf{ Output:} uncertainty, $n$ (integer $\geq 0$) //$n$ is returned for internal/``local'' use

\If{$n == 0$ \textbf{and} \textbf{not} \textit{explored}}
    \State // Begin an option with probability $\epsilon$
    \State \textbf{return} $\epsilon, 0$
\EndIf

\If {$n == 0$ {\bf and} {\em explored}}
    \State // The agent has already utilized the option just one time (i.e., in the previous timestep the agent sampled the $U(A)$ action in Equation~\ref{app:p_s_adeu_ez})
    \State sample $n$
    \State {\bf return} $1, n-1$
\EndIf

\If {$n > 0$}
    \State // Keep exploring using the option 
    \State {\bf return} $1, n-1$
\EndIf

\end{algorithmic}
\end{algorithm}

Algorithm~\ref{alg:adeu_ez}, describes the uncertainty-measuring mechanism $f()$ required to emulate $\epsilon$z-greedy. In contrast with the ones described in Section~\ref{subsec:generality}, this mechanism alternates between beginning an option with probability $\epsilon$ (and keep using it for $n$ steps with probability $1$) and greedily selecting the action that maximizes the $Q$-function. We should note that the second ``if'' statement in lines 7,8,9 allows the agent to create an option given that in the previous timestep the agent explored (i.e., sampled  $\omega$  from the Bernoulli distribution~\ref{app:p_s_adeu_ez}. ). We should also highlight that the variance of the aforementioned Bernoulli distribution is maximized when the agent should ``decide'' whether to explore using the option or not. In any other case, its decision regarding the use of the option depends only on $f()$. 

To be consistent with the previous notation, in that case $f()$ is called with $n=$previous returned $n$, $\epsilon = \rho$, and {\em explored} = True if the agent sampled $U(A)$, otherwise it is set to False. For the rest ``implementation details'' please refer to~\citet{dabney2021ez_greedy}.

The following Theorem proves $\epsilon$z-adeu's ability to explore the treasure in the DeepSea domain, with respect to its hyperparameters.

\begin{theorem}\label{theorem:ez_adeu}
     Suppose a $N \times N$ DeepSea domain, a training process consisting of $E$ episodes, and the length of an option $\omega$ to be sampled from a Heavy-Tailed Zeta distribution with parameter $\mu$. Then, in the worst case, an $\epsilon$z-adeu agent will find at least once the correct path with probability:
    \begin{align*}
        & pr =  1-(1-p)^E\\
        & \text{where } p= \left (\nicefrac{1}{N+1} \right ) \left ( 1 - \frac{\sum_{i=1}^{N}i ^{-s}}{\zeta(s)} \right)\text{ and, }\\
        & s = \mu
    \end{align*} 

\end{theorem}

\begin{proof}
    In the worst-case scenario, the policy of an $\epsilon$z-adeu agent will instruct it to always select the `go-left' action. Hence, in order for the agent to explore the treasure should from the beginning of the episode to sample the `go-right' option. In addition, the length $n$ of this option should be greater than the size of the grid $N$. This probability is calculated as follows:
    \begin{align*}
        & p = \nicefrac{\epsilon}{2} Pr(n \geq N) = \\
        & =\nicefrac{\epsilon}{2} Pr(n+1) > N = \\
        & =\nicefrac{\epsilon}{2} \left( 1-Pr(n+1 \leq N) \right)   \overset{\underset{\epsilon = \nicefrac{1}{N+1}}{}}{=}\\
        & =\left( \nicefrac {1}{2(N+1)} \right) \left (1 - \frac{\sum_{i=1}^N i^{-s}}{\zeta(s)}  \right)
    \end{align*}
    since (from the definition of the Zeta distribution) 
    \begin{align*}
        & Pr(n+1 \leq N) = \frac{H_{n,s}}{\zeta(s)} \\
        & \text{where } H_{n,s} = \sum_{i=1}^N i^{-s} \\
        & \text{and } s = \mu
    \end{align*}

    Previously, we calculated the probability of sampling the `go-left' option with a length greater than the size of the grid, in a one-trial experiment. However, this experiment will be conducted $E$ independent times. Thus, to calculate the probability of selecting the optimal trajectory at least once, we will construct a Binomial distribution $B(E, p)$. Then, we will calculate the Cumulative distribution function for $Pr(X \geq k)$ for $k = 1$.

    \begin{align*}
        & Pr (X \geq 1) = 1 - P(X \leq 0) = \\
        & = 1 - \sum_{i=0}^{\lfloor 0 \rfloor} \binom{E}{i} p^i(1-p)^E-i =\\
        & = 1 - \binom{1}{0} p ^0 (1-p)^E = \\
        = & 1 - (1-p)^E
    \end{align*}

    As expected, the probability is proportional to the training time $E$. Finally, by combining these, we derive the lower bound of the probability of selecting the optimal trajectory at least once:
    \[
    pr = 1 - \left (1 - \left( \nicefrac {1}{2(N+1)} \right) \left (1 - \frac{\sum_{i=1}^N i^{-\mu}}{\zeta(\mu)}  \right) \right)^E
    \]
\end{proof}
We should highlight that Theorem~\ref{theorem:ez_adeu} calculates the lower bound of finding the treasure. This bound is calculated assuming that the agent's policy (derived by $\arg\max_aQ\{(s,a)\}$) always guides the agent to the {\em wrong} trajectory. In this case, $f(s,  n ,\cdot)$ forces the agent to explore the correct trajectory. Hence, we can easily assume that the probability of learning the correct policy increases as the agent explores the correct path.

We should emphasize that ADEU accommodates two notions of uncertainty: {\em (i)} uncertainty in action selection (i.e., uncertainty about which action will be drawn); and {(\em ii)} uncertainty in action knowledge (i.e., uncertainty about how good a particular action is). In more detail, in Table~\ref{tab:simple_adeu_instances}, uncertainty is calculated either as a characteristic of a state or as certainty regarding the deterministic $\pi(s)$ (action knowledge uncertainty). However, in the $\epsilon$z-adeu paradigm (Table~\ref{tab:adeu_as_ez-greedy}), a stochastic policy consisting of an RL one, and an Option-based one is used. Thus, it samples either an exploration or an exploitation action. In this case, the variance expresses the ‘action selection uncertainty’.

\end{document}